\documentclass[10pt,twocolumn,letterpaper]{article}

\usepackage{cvpr}
\usepackage{times}
\usepackage{epsfig}
\usepackage{graphicx}
\usepackage{amsmath}
\usepackage{amssymb}
\usepackage{xcolor}
\usepackage{textcomp}
\usepackage{color}
\usepackage{colortbl}
\usepackage{xfrac}

\usepackage{float}

\usepackage{caption}
\usepackage{subcaption}
\usepackage{stackengine}
\usepackage{tabularx}

\usepackage{paralist}  
\usepackage{mathrsfs}  
\usepackage{multirow}

\definecolor{yellow}{rgb}{1,1,0.8}
\definecolor{orange}{rgb}{1,0.8, 0.6}
\definecolor{red}{rgb}{1, 0.6, 0.6}
\definecolor{darkred}{rgb}{0.8, 0.1, 0.1}
\definecolor{blue}{rgb}{0, 0, 1.0}
\definecolor{green}{rgb}{0, 1.0, 0}
\definecolor{magenta}{rgb}{1.0, 0, 1.0}
\definecolor{white}{rgb}{1.0, 1.0, 1.0}

\newcommand{\RGBA}{RGB$\alpha$~} 
\newcommand{\level}{\ell} 

\newcommand{\lossfun}[1]{\mathcal{L}_\mathit{#1}}  

\usepackage[pagebackref=true,breaklinks=true,letterpaper=true,colorlinks,bookmarks=false]{hyperref}

\cvprfinalcopy 

\def\cvprPaperID{6586} 

\ifcvprfinal\fi
\begin{document}

\title{Lighthouse: Predicting Lighting Volumes for Spatially-Coherent Illumination}

\author{Pratul P. Srinivasan*$^1$
\qquad
Ben Mildenhall*$^1$
\qquad
Matthew Tancik$^1$ \\
Jonathan T. Barron$^2$
\qquad
Richard Tucker$^2$
\qquad
Noah Snavely$^2$ \\
$^1$UC Berkeley, $^2$Google Research
}

\maketitle
\thispagestyle{empty}

\begin{abstract}

We present a deep learning solution for estimating the incident illumination at any 3D location within a scene from an input narrow-baseline stereo image pair. Previous approaches for predicting global illumination from images either predict just a single illumination for the entire scene, or separately estimate the illumination at each 3D location without enforcing that the predictions are consistent with the same 3D scene. Instead, we propose a deep learning model that estimates a 3D volumetric \RGBA model of a scene, including content outside the observed field of view, and then uses standard volume rendering to estimate the incident illumination at any 3D location within that volume. Our model is trained without any ground truth 3D data and only requires a held-out perspective view near the input stereo pair and a spherical panorama taken within each scene as supervision, as opposed to prior methods for spatially-varying lighting estimation, which require ground truth scene geometry for training. We demonstrate that our method can predict consistent spatially-varying lighting that is convincing enough to plausibly relight and insert highly specular virtual objects into real images.

\end{abstract}

\begin{figure}[!]
\begin{center}
\newcommand{\width}{1.0\linewidth}
\includegraphics[width=\width]{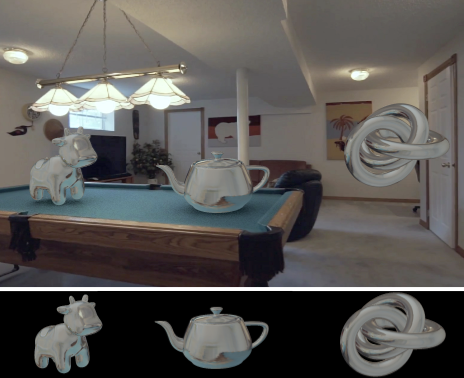}
\caption{Our method predicts environment map lighting at any location in a 3D scene from a narrow-baseline stereo pair of images. We use this to convincingly insert specular objects into real photographs with spatially-coherent lighting that varies smoothly in 3D. Below, we isolate the relit objects to better visualize our estimated illumination. Notice how each inserted object contains different specular highlights and reflected colors corresponding to its 3D location, such as the light reflected on the cow's head and the corner of the table visible on the teapot.}
\label{fig:teaser}
\end{center}
\end{figure}

\begin{figure*}
\begin{center}
\newcommand{\width}{1.0\linewidth}
\includegraphics[width=\width]{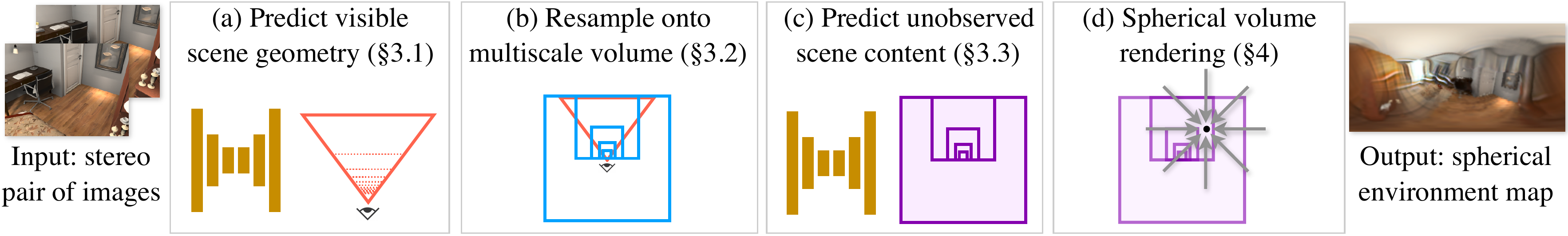}
\caption{Our method takes a narrow-baseline stereo pair of images as input, uses a 3D CNN to predict an intermediate representation of visible scene geometry (a), resamples this onto a multiscale volume that encompasses unobserved regions of the scene (b), completes this volume with another 3D CNN (c), and renders spatially-coherent environment maps at any 3D location from this same volume using standard volume tracing (d).}
\label{fig:pipeline}
\end{center}
\end{figure*}

\let\thefootnote\relax\footnote{* Authors contributed equally to this work.}

\section{Introduction}

Rendering virtual objects into photographs of real scenes is a common task in mixed reality and image editing. Convincingly inserting such objects requires estimating both the geometry of the scene (so that inserted objects are correctly occluded by real scene content) as well as the incident illumination at points on each object's surface (so that inserted objects appear to be lit by the surrounding environment).
The difficulty of this task is exacerbated by the fact that incident illumination can vary significantly across different locations within a scene, especially indoors, due to lights close to the inserted objects, shadows cast by scene geometry, and global illumination effects from nearby scene content.
Additionally, compositing large objects, multiple objects, or objects that move within the scene is even more difficult as doing so requires estimating a spatially-varying model of illumination that is \emph{spatially-coherent} (we (ab)use this term to mean that it varies smoothly as a function of position in accordance with a plausible 3D scene).

Current state-of-the-art algorithms for estimating global illumination either predict a single illumination for the entire scene \cite{calian2018face2light,gardner-sigasia-17,holdgeoffroy-cvpr-17,lalonde-iccv-09,legendre2019deeplight,sengupta19} or estimate spatially-varying illumination by separately predicting the lighting at individual 3D locations within the scene \cite{garon2019,li19, song2019neural}.
The single-illumination approach can only be used to illuminate small objects at a predefined location, while the separately-predicted spatially-varying approach can produce compelling results, but does not guarantee that the predicted illumination will vary smoothly as a function of position.
In this work, we propose an algorithm that predicts a volumetric representation of the scene from a narrow-baseline stereo pair of images, and then uses that volumetric representation to produce a spatially-varying model of illumination by simply rendering that predicted volume from the set of required object insertion locations.
Because our approach computes each environment map from a single predicted underlying volumetric scene representation using standard volume rendering, all estimated lighting is naturally consistent with the same 3D scene, and lighting locations can be queried at 100 frames per second allowing for real-time object insertion.

We specifically use a narrow-baseline stereo pair as input because: 1) multi-camera systems are ubiquitous in modern smartphones, 2) stereo enables us to estimate the high fidelity geometry required for simulating spatially-varying lighting effects due to observed scene content, and 3) we can leverage recent progress in using narrow-baseline stereo images to predict 3D scene representations for view synthesis~\cite{SrinivasanCVPR2019,zhou2018stereo}, enabling us to render novel views of the scene with relit objects for virtual reality object insertion.

To summarize, our primary technical contributions are:
\begin{enumerate}
    \item A multiscale volumetric scene lighting representation that is specifically designed for estimating realistic spatially-varying lighting (Sec.~\ref{sec:resampling}) and a deep learning--based approach for predicting this representation using only a narrow-baseline stereo pair as input (Sec.~\ref{sec:completion}). We design this representation to support rendering spatially-varying illumination without any network inference (Sec.~\ref{sec:rendering}), so lighting prediction is very fast and guaranteed to be spatially-coherent.
    \item A training procedure that only needs perspective and panoramic views of scenes for supervision, instead of any ground-truth 3D scene representation (Sec.~\ref{sec:training}).
\end{enumerate}
We demonstrate that estimating spatially-varying global illumination as a persistent 3D function quantitatively and qualitatively outperforms prior approaches. Our spatially-coherent estimated lighting can simulate convincing global illumination effects for rendering specular virtual objects moving within scenes. We encourage readers to view animations of these results in our supplementary video.

\section{Related Work}

\subsection{Estimating lighting from images}

Inferring the intrinsic properties of lighting, materials, and geometry that together form an image is a fundamental problem that has been studied in various forms throughout the history of computer vision~\cite{Barrow1978,Horn1974DeterminingLF}. Below, we review relevant prior works that use images to estimate representations of lighting for relighting virtual objects.

Seminal work by Debevec~\cite{Debevec2008} showed that virtual objects can be convincingly inserted into real photographs by rendering virtual object models with high dynamic range (HDR) environment maps captured with bracketed exposures of a chrome ball. Many subsequent methods~\cite{calian2018face2light,gardner-sigasia-17,holdgeoffroy-cvpr-17,lalonde-iccv-09,legendre2019deeplight,sengupta19} have demonstrated that machine learning techniques can be used to estimate an HDR environment map from a single low dynamic range (LDR) photograph.

However, a single environment map is insufficient for compositing multiple, large, or moving virtual objects into a captured scene, especially in indoor settings where light sources and other scene content may be close to the object insertion locations. To address this shortcoming, many works predict spatially-varying lighting from images by estimating a separate environment map for each pixel in the input image. Such approaches include algorithms designed specifically for spatially-varying lighting estimation~\cite{garon2019} as well as methods that address the more general inverse rendering problem of jointly estimating the spatially-varying lighting, scene materials, and geometry that together produce an observed image~\cite{li19,Shelhamer2015}. However, these approaches do not ensure that the illuminations predicted at different spatial locations correspond to a single 3D scene, and their approach of indexing lighting by image pixel coordinates cannot estimate lighting at locations other than points lying directly on visible scene surfaces. Karsch \etal~\cite{Karsch2014} also address a similar inverse rendering problem, but instead estimate area lights in 3D by detecting visible light source locations and retrieving unobserved light sources from an annotated panorama database.

Our work is closely related to prior deep learning methods that estimate a portion of lighting in 3D.
Neural Illumination~\cite{song2019neural} predicts the incident illumination at a location by first estimating per-pixel 3D geometry for the input image, reprojecting input image pixels into an environment map at the queried location, and finally using a 2D CNN to predict unobserved content in the resulting environment map.
This strategy ensures spatial consistency for light emitted from scene points that are visible in the input image (for which a single persistent geometry estimate is used), but because the environment map is separately completed for each lighting location using a 2D CNN, the lighting from unobserved scene points is not spatially-coherent.
Recent work by Gardner \etal~\cite{gardner2019} trains a deep network to estimate the positions, intensities, and colors of a fixed number of light sources in 3D, along with an ambient light color. This ensures spatially-coherent lighting, but is unable to simulate realistic global illumination effects or light source occlusions, which can be very significant in indoor scenes, and therefore has difficulty rendering realistic specular objects.
Furthermore, both of these methods require ground truth scene depths for training while our method only requires perspective and spherical panorama images.

\subsection{Predicting 3D scene representations}

Our strategy of estimating consistent spatially-varying lighting by predicting and rendering from a 3D scene representation is inspired by recent successes in using 3D representations for the image-based rendering problem of predicting novel views of a scene. Shum and Kang~\cite{Shum2006} provide an excellent review of classic approaches, ranging from light field rendering methods~\cite{levoy1996light} that do not use any scene geometry, to texture mapping methods~\cite{Debevec1996,Shade1998} that use a global scene mesh. A key lesson from early work on image-based rendering is that more knowledge of scene geometry reduces the number of sampled images required for rendering new views~\cite{buehler01,chai00}.
Modern approaches to view synthesis follow this lesson, rendering novel views by predicting representations of 3D scene geometry from sparsely-sampled collections of images. In particular, many recent methods predict layered or volumetric 3D scene representations, which have a regular grid structure that is well-suited to CNN pipelines. This includes algorithms for synthesizing novel outwards-facing views of large scenes~\cite{flynn16,mildenhall2019llff,SrinivasanCVPR2019,zhou2018stereo} and inwards-facing views of objects~\cite{Lombardi2019,sitzmann2019deepvoxels,xu19}.

We adopt the approach of Zhou \etal~\cite{zhou2018stereo} for learning to predict a layered representation of observed scene content. Their algorithm trains a CNN to predict a set of fronto-parallel \RGBA planes sampled evenly in disparity within the camera frustum. Training proceeds by minimizing the difference between renderings of their model and held-out novel views, thereby obviating the need for ground truth 3D supervision. This representation, which they call a multiplane image (MPI), is closely related to representations used in volume rendering~\cite{drebin1988volume,lacroute1994,Levoy1988} and stereo matching~\cite{Szeliski1999}.
Though this approach works well for view synthesis, it cannot be directly used for estimating spatially-varying lighting, as the majority of the illumination needed for relighting and compositing virtual objects often resides \emph{outside} of the input image's field of view.
We address this shortcoming by extending these models to predict a multiscale volumetric representation that includes scene content outside the input image's camera frustum, thereby allowing us to estimate incident illumination at any 3D location in the scene.

\section{Multiscale Lighting Volume Prediction}
\label{sec:volume}

Our goal is to take in a narrow-baseline pair of RGB images and associated camera poses, and output the incident illumination (represented as a spherical environment map) at any queried 3D location within the scene.
Due to dataset limitations (see Sec.~\ref{sec:datasetdetails}), we do not address LDR-to-HDR conversion and assume that the inputs are either HDR captures or that they can be converted to HDR by inverting a known tone mapping curve or by applying existing LDR-to-HDR conversion techniques~\cite{eilertsen2017hdr}.

We train a deep learning pipeline, visualized in Fig.~\ref{fig:pipeline}, that regresses from the input image pair to a volumetric \RGBA representation of the entire scene that includes areas outside of the reference camera frustum (we choose one of the two input images as the ``reference'' to be the center of our coordinate system), thereby allowing the illumination at any 3D location to be estimated by simply rendering the scene volume at that location. This representation enables us to reproduce effects such as: shadowing due to the occlusion of light sources by other scene content, realistic reflections on glossy and specular virtual objects, and color bleeding from the scene onto relit objects.

\begin{figure}
\begin{center}
\newcommand{\width}{1.0\linewidth}
\includegraphics[width=\width]{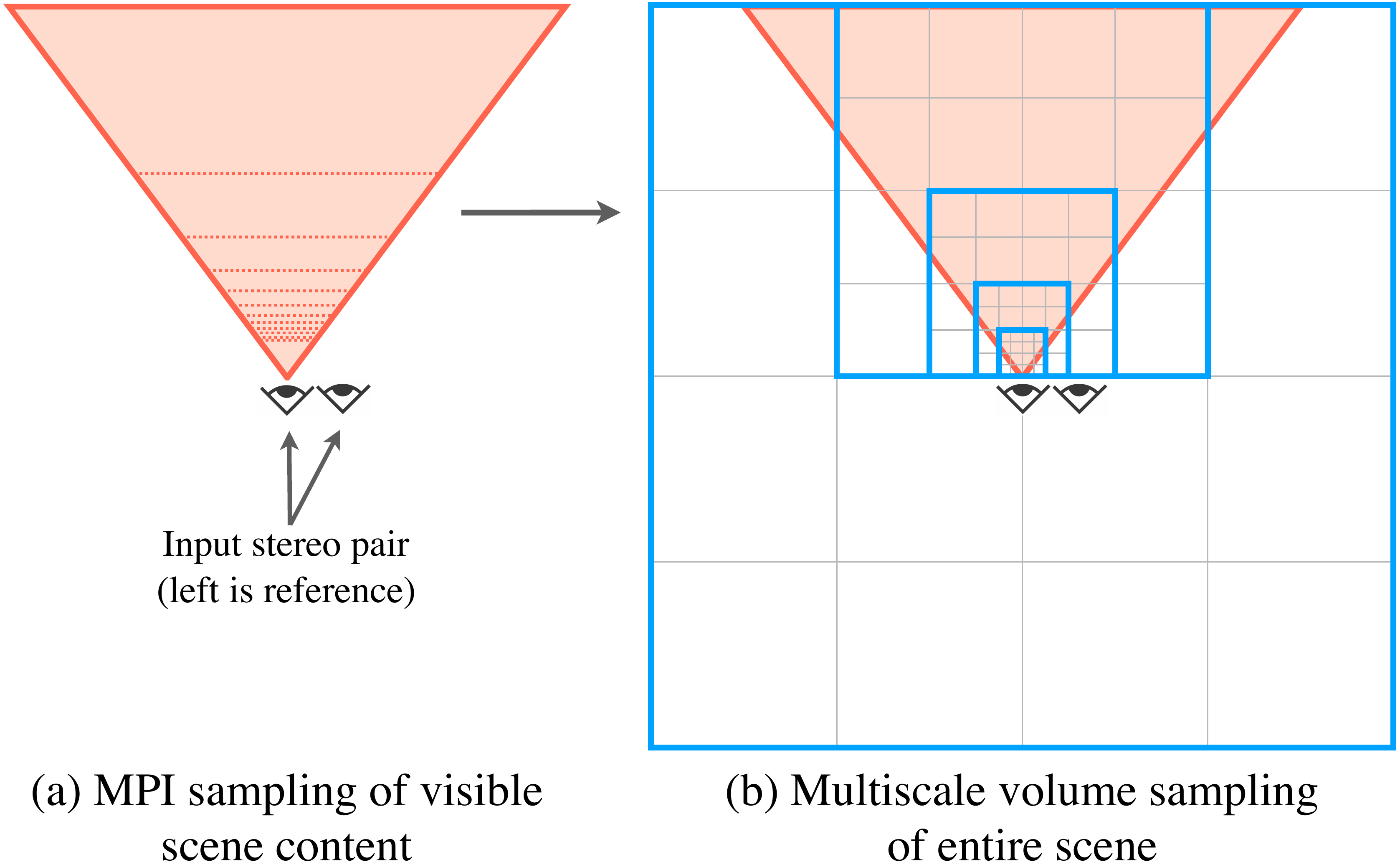}
\caption{A 2D visualization of our 3D multiscale volume resampling. (a) First, given an input stereo pair of images, we predict scene content within the reference camera frustum as a set of \RGBA planes spaced linearly in inverse depth. (b)~Next, we resample the frustum geometry onto a set of nested cubes with increasingly finer sampling, centered around the input camera.}
\label{fig:multiscale_resampling}
\end{center}
\end{figure}

Na\"ively representing an entire indoor scene as a dense high-resolution voxel grid is intractable due to memory constraints. Instead, we propose a multiscale volume representation designed to adequately sample the varying depth resolution provided by stereo matching, allocate sufficient resolution to areas where virtual objects would be inserted, and allocate lower resolution outside the observed field-of-view where scene content must be hallucinated.
As shown in Fig.~\ref{fig:pipeline} our procedure for predicting this volume is: 1) A deep network predicts a layered representation of observed scene content from the input narrow-baseline stereo pair (Fig.~\ref{fig:pipeline}a). 2) This layered representation is resampled onto a multiscale lighting volume that preserves the observed content representation's resolution (Fig.~\ref{fig:pipeline}b). 3) Another deep network completes the multiscale lighting volume by hallucinating scene geometry and appearance outside the observed field of view (Fig.~\ref{fig:pipeline}c). Once we generate a lighting volume for a scene, we can use classic volume rendering to estimate the incident illumination at any 3D location without requiring any network inference (Fig.~\ref{fig:pipeline}d).

\subsection{Observed content intermediate representation}
\label{sec:mpi}

We construct an intermediate representation of observed scene content as an MPI $M$, which consists of a set of fronto-parallel \RGBA planes within the frustum of a reference camera, as visualized in Fig.~\ref{fig:multiscale_resampling}a. As in Zhou \etal~\cite{zhou2018stereo}, we select one of the input images as a reference view and construct a plane sweep volume (PSV) in this frame for both input images. We concatenate these two PSVs along the channel dimension, forming an input tensor for a 3D encoder-decoder CNN that outputs an MPI, as suggested by follow-up works on MPI prediction~\cite{mildenhall2019llff,SrinivasanCVPR2019}.

\subsection{Multiscale volume resampling}
\label{sec:resampling}

This MPI provides an estimation of geometry for regions of the scene observed in the two input images. However, inserting a relit virtual object into the scene also requires estimating geometry and appearance for unobserved areas behind and to the sides of the input cameras' frustums, as visualized in Fig.~\ref{fig:multiscale_resampling}, so our lighting representation must encompass this area. Furthermore, our volumetric lighting representation should allocate higher resolution to regions where we would insert objects in order to correctly render the larger movement of nearby content within environment maps as the queried location changes.

We design a multiscale volume lighting representation that encompasses both observed and unobserved regions, allocates finer resolution within the input field-of-view, and increases in resolution towards the front of the MPI frustum (which contains increasingly higher resolution estimated geometry from stereo matching). We initialize this multiscale volume by resampling the \RGBA values from the MPI frustum onto a series of nested cubes $V^o = \{V^o_1, \ldots, V^o_L\}$ encompassing the whole scene, using trilinear interpolation. From the coarsest to finest level, each cube $V^o_\level$ is half the spatial width of the previous level $V^o_{\level-1}$ while maintaining the same grid resolution of $64^3$. The largest, outermost cube $V^o_1$ is centered at the first input camera pose and is wide enough to contain the whole MPI volume. Each smaller nested cube is offset such that the input camera pose lies at the back face of the cube. See Fig.~\ref{fig:multiscale_resampling} for a visualization of the MPI sampling pattern and how we resample it onto the multiscale volume structure. We find that this multiscale sampling pattern works well in practice for rendering convincing near-field lighting effects caused by scene geometry at our chosen environment map resolution of $120\times240$.

\begin{figure}
\begin{center}
\newcommand{\width}{0.8\linewidth}
\includegraphics[width=\linewidth]{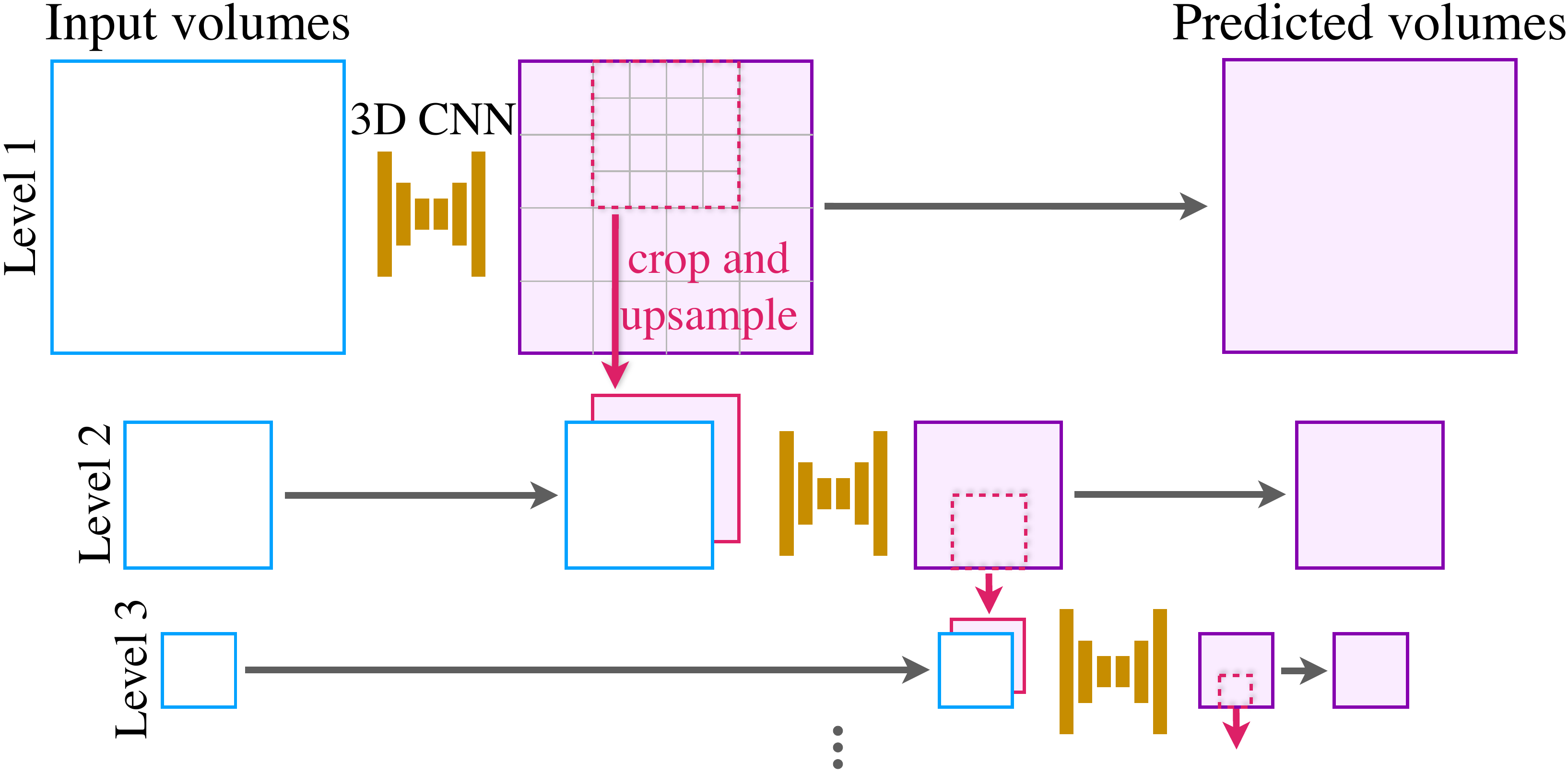}
\caption{A visualization of our multiscale volume completion network. After resampling the visible scene geometry onto a series of nested cubes, we apply a volume completion network to hallucinate the unseen geometry at each level. The prediction is done in a coarse-to-fine manner, where the coarse prediction at level $\level$ is cropped and upsampled then fed into the CNN along with the resampled visible volume to predict level $\level+1$.}
\label{fig:multiscale_network}
\end{center}
\end{figure}

\subsection{Multiscale volume completion}
\label{sec:completion}

Now that we have a volumetric representation $V^o = \{V^o_1, \ldots, V^o_L\}$ of the entire scene that has been populated with observed content, we use a deep network to hallucinate the geometry and appearance of the unobserved content. We denote this ``completed'' multiscale volume by $V^c = \{V^c_1, \ldots, V^c_L\}$. We design a 3D CNN architecture that sequentially processes this multiscale volume from the coarsest to the finest resolution, predicting a completed volume $V^c_\level$ at each resolution level. For each level, we first nearest-neighbor upsample the region of the previous coarser completed volume $V^c_{\level-1}$ that overlaps the current level $V^o_\level$ to $64^3$ resolution.
Then, we concatenate this to the current level's resampled volume $V^o_\level$ along the channel dimension and use a 3D encoder-decoder CNN to predict the current level's completed volume $V^c_\level$, with separate weights for each level. Figure~\ref{fig:multiscale_network} visualizes our coarse-to-fine network architecture. Please refer to our supplementary PDF for exact architecture specifications.

\begin{figure}
\begin{center}
\newcommand{\width}{1.0\linewidth}
\includegraphics[width=\width]{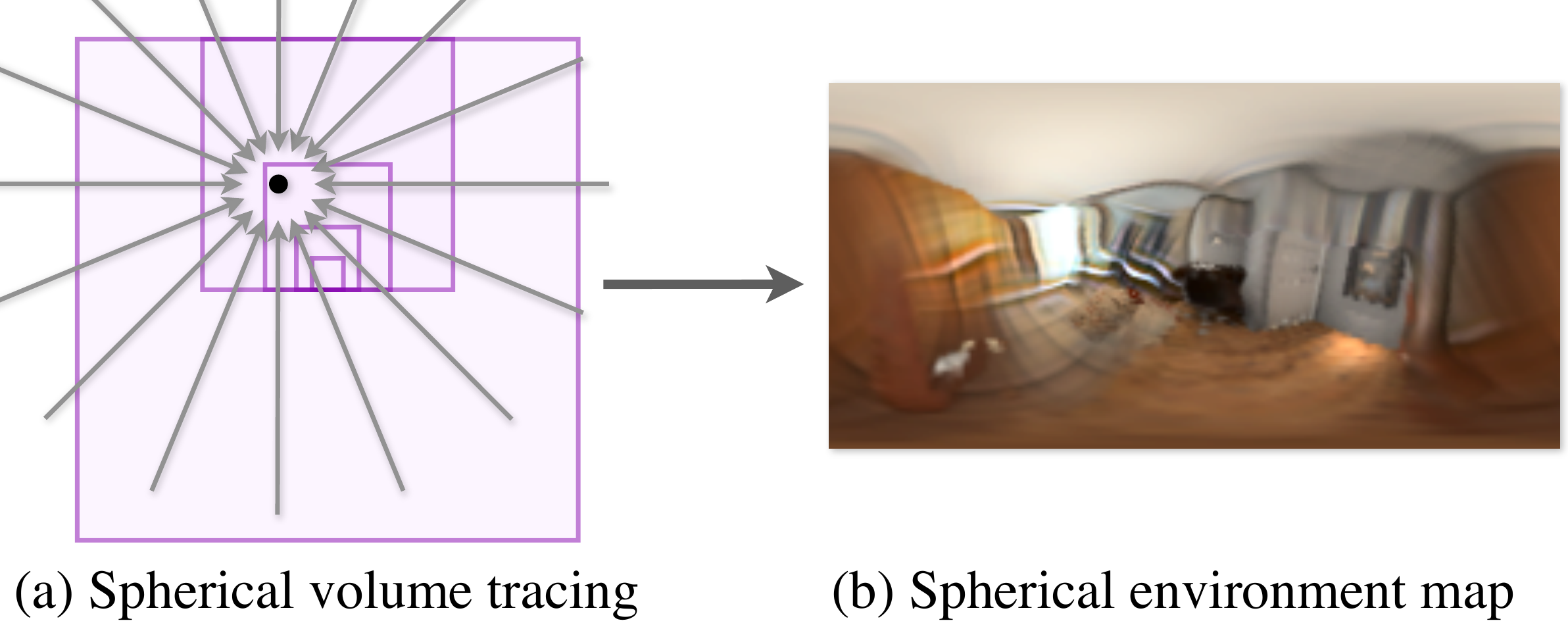}
\caption{Given our predicted multiscale lighting volume and a 3D location in the scene (shown as a black circle), we render the environment map by (a) tracing spherical rays through the volumes and alpha compositing from the outermost to the innermost \RGBA value, producing a single spherical environment map (b).}
\label{fig:cube_rendering}
\end{center}
\end{figure}

\section{Illumination Rendering}
\label{sec:rendering}

Given our multiscale lighting volume $V^c$, we estimate the illumination incident at any 3D location by using standard \RGBA volume rendering~\cite{Levoy1988} to generate a spherical environment map at that point (visualized in Fig.~\ref{fig:cube_rendering}). In order to get the value at a pixel $p$ for an environment map located at location $\mathbf x$, we must:
\begin{enumerate}
    \item Generate the ray $r$ (in world coordinates) that originates at $\mathbf x$ and intersects pixel $p$ on the sphere, and
    \item Trace $r$ through the volume $V^c$, using alpha compositing to matte in the \RGBA values as it intersects voxels from farthest to nearest.
\end{enumerate}
As we trace $r$ through $V^c$, we query the finest level defined at that location in space to ensure that predicted \RGBA values at coarser levels never override predictions at finer levels. This rendering procedure is very fast since it does not involve network inference and is trivially parallelizable on GPUs, allowing us to render environment maps from a predicted multiscale volume at 100 frames per second.

\section{Training and Dataset}
\label{sec:training}

Our model is trained end-to-end: a stereo pair is provided as input, the model renders a held-out novel view (sampled close to the reference view) from the intermediate MPI and a held-out environment map (sampled within the scene in front of the reference camera) from the completed multiscale lighting volume, and we update the model parameters only using the gradient of losses based on these two supervision images. This is possible since all steps in our pipeline are differentiable, including the multiscale volume resampling and the environment map rendering. Therefore, we do not require ground-truth geometry or other labels, in contrast to prior works in spatially-varying lighting estimation which either require scene geometry as supervision~\cite{garon2019,li19,sengupta19,song2019neural} or for creating training data~\cite{gardner2019}.

\subsection{Training loss}

The loss we minimize during training is the sum of an image reconstruction loss for rendering a held-out perspective view from our predicted MPI, an image reconstruction loss for rendering an environment map from our completed multiscale lighting volume, and an adversarial loss on the rendered environment map to encourage plausible high frequency content.
For our reconstruction loss, we use a perceptual loss $\lossfun{vgg}$ based on features from a pre-trained VGG-19 network~\cite{SimonyanZ14a}, as done by Chen and Koltun~\cite{chen2017photographic}.
For our adversarial loss, we follow recent work in conditional image generation~\cite{park19,wang18} and use a PatchGAN~\cite{isola17} discriminator $\mathcal{D}$ with spectral normalization~\cite{miyato18} and a hinge adversarial loss~\cite{lim17}.
We train all networks in our pipeline by alternating between minimizing the reconstruction and adversarial losses with respect to the MPI prediction and volume completion networks' parameters:
\begin{equation}
    \lossfun{train} = \lossfun{vgg}(i_{r},i_{gt}) + \lossfun{vgg}(e_{r},e_{gt})-\mathcal{D}(e_{r}),
\end{equation}
and minimizing the discriminator loss with respect to the discriminator network's parameters:
\begin{equation}
    \lossfun{dis} = \max\left(0, 1 - \mathcal{D}\left(e_{gt}\right)\right) + \max\left(0, 1 + \mathcal{D}\left(e_{r}\right)\right),
\end{equation}
where $i_{r},i_{gt},e_{r},e_{gt}$ are the rendered and ground truth perspective image and environment maps, respectively.

\subsection{Dataset details}
\label{sec:datasetdetails}

We train our model with photorealistic renderings of indoor scenes from the InteriorNet dataset~\cite{interiornet2018}. We use 1634 of the provided camera sequences, each containing 1000 perspective projection images and 1000 spherical panorama images rendered along the same camera path. We reserve 10\% of these sequences for our test set, and sample training examples from the remaining 90\% of the sequences.

The images included in InteriorNet are not HDR, and unfortunately no equivalent dataset with HDR radiance values currently exists. In our experiments, we assume that the InteriorNet images can be treated as linear radiance values by applying an inverse gamma curve $x^\gamma$, with $\gamma=2.2$.

To generate training examples from a sequence, we randomly sample three perspective projection images, evenly separated with a gap between 1 and 8 frames along the sequence's camera path, and a single spherical panorama image within 40 frames of the central perspective projection frame. Two of the perspective projection images are randomly selected to be the input to our model, while the third perspective image and the spherical panorama image are used as supervision. We reject training examples where the camera is closer than 0.1m from the scene, examples where adjacent perspective cameras are separated by less than 0.05m, and examples where the average pixel brightness is lower than 0.1. Additionally, we reject examples where the spherical panorama camera does not move more than the median scene depth into the scene, relative to the central perspective projection camera, so that the environment map locations we use for supervision are representative of realistic object insertion locations.

\begin{table}
\begin{center}
\resizebox{\linewidth}{!}{
\begin{tabular}{@{}c@{\,\,}l|cc}
& Method & PSNR (dB) $\uparrow$  & Angular Error (\textsuperscript{$\circ$}) $\downarrow$ \\
\hline
\multirow{6}{*}{\rotatebox{90}{all content}} &DeepLight~\cite{legendre2019deeplight} & $13.36 \pm 1.29 $  & $7.15 \pm 3.24 $   \\
&Garon \etal~\cite{garon2019} & $13.21 \pm 1.80 $  & $12.73 \pm 7.17 $   \\
&Neural Illumination~\cite{song2019neural} & $16.59 \pm 1.91 $  & $5.26 \pm 2.84 $   \\
&Ours (MPI only) & $15.26 \pm 2.11 $  & $6.41 \pm 3.42 $    \\
&Ours (no $\lossfun{adv}$) & \cellcolor{yellow} $17.54 \pm 1.97 $  & $4.74 \pm 2.70 $   \\
&Ours &  $17.29 \pm 1.97 $  & \cellcolor{yellow} $4.71 \pm 2.68 $
\vspace{0.1in} \\
\multirow{6}{*}{\rotatebox{90}{observed content}} &DeepLight~\cite{legendre2019deeplight} & $13.94 \pm 1.96 $  & $7.21 \pm 4.05 $   \\
&Garon \etal~\cite{garon2019} & $14.52 \pm 2.30 $  & $12.33 \pm 9.03 $   \\
&Neural Illumination~\cite{song2019neural} & $18.58 \pm 3.55 $  & $4.97 \pm 3.42 $   \\
&Ours (MPI only) & $17.97 \pm 3.86 $  & $4.45 \pm 4.46 $    \\
&Ours (no $\lossfun{adv}$) & $19.74 \pm 3.64 $  & $4.18 \pm 3.29 $   \\
&Ours & \cellcolor{yellow} $19.79 \pm 3.99 $  & \cellcolor{yellow} $3.76 \pm 3.09 $   \\
\end{tabular}
}
\caption{Quantitative results for rendered environment maps. We separately report performance on all content (the complete environment map) and on only observed content (the portion of each environment map that was observed in the input image). We report the PSNR and RGB angular error (following~\cite{legendre2019deeplight}) for the predictions for each method versus ground truth spherical environment maps.
}
\label{tab:comparisons_both}
\end{center}
\end{table}

\subsection{Additional details}

We implement our full training pipeline in TensorFlow~\cite{tensorflow} and train our model on a single NVIDIA Tesla V100 GPU using the Adam optimizer~\cite{adam} with a batch size of 1.
For more stable training, we first pre-train the MPI prediction network for 240k iterations, then train both the MPI prediction network and volume completion networks with just image reconstruction losses for 450k iterations, and finally add the adversarial losses and train both networks along with a discriminator for an additional 30k iterations. We use an Adam step size of $10^{-4}$ for the first two stages and $10^{-5}$ for the third stage.

\section{Results}

We validate the benefits of our algorithm by comparing our estimated environment maps to those of current state-of-the-art algorithms and ablated versions of our model. Please view our supplementary video for example results that demonstrate our method's ability to estimate realistic and consistent spatially-varying lighting.

For quantitative comparisons (Table~\ref{tab:comparisons_both}), we sample a test set of 4950 examples (using the same training example rejection criteria described above in Sec.~\ref{sec:datasetdetails}) from our InteriorNet test set, which consists of 163 camera sequences that were held out during training. Each example consists of two input images and one ground truth environment map that represents the lighting at a random 3D location within the reference camera frustum. We select a subset of these examples to show comparisons of virtual object insertion results in Fig.~\ref{fig:comparisons} and show additional insertion results for our method on real photographs in Fig.~\ref{fig:realestate_insertion}.

\newcommand{\eximgwidth}{.162\textwidth}
\newcommand{\metricspace}{\,\,}

\begin{figure*}
\captionsetup[subfigure]{font=scriptsize,labelformat=empty,aboveskip=1pt,belowskip=2pt}
  \centering

  \begin{subfigure}[t]{\eximgwidth}
      \centering Ref. Image
  \end{subfigure}
  \begin{subfigure}[t]{\eximgwidth}
      \centering Ground Truth
  \end{subfigure}
  \begin{subfigure}[t]{\eximgwidth}
      \centering DeepLight~\cite{legendre2019deeplight}
  \end{subfigure}
  \begin{subfigure}[t]{\eximgwidth}
      \centering Garon \etal~\cite{garon2019}
  \end{subfigure}
  \begin{subfigure}[t]{\eximgwidth}
      \centering Neural Illum.~\cite{song2019neural}
  \end{subfigure}
  \begin{subfigure}[t]{\eximgwidth}
      \centering Ours
  \end{subfigure}

  \vspace{0.0in}
  \begin{subfigure}[t]{\eximgwidth}
      \centering\includegraphics[width=\textwidth]{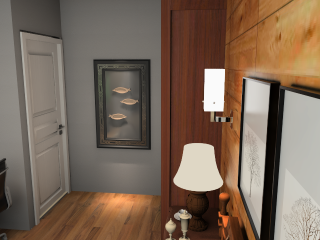}
  \end{subfigure}
  \begin{subfigure}[t]{\eximgwidth}
      \centering\includegraphics[width=\textwidth]{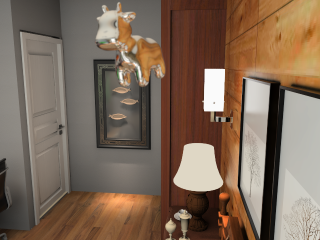}
  \end{subfigure}
  \begin{subfigure}[t]{\eximgwidth}
      \centering\includegraphics[width=\textwidth]{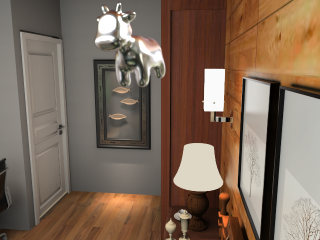}
  \end{subfigure}
  \begin{subfigure}[t]{\eximgwidth}
      \centering\includegraphics[width=\textwidth]{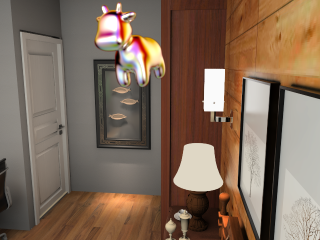}
  \end{subfigure}
  \begin{subfigure}[t]{\eximgwidth}
      \centering\includegraphics[width=\textwidth]{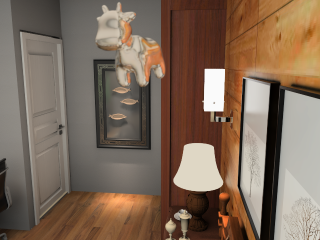}
  \end{subfigure}
  \begin{subfigure}[t]{\eximgwidth}
      \centering\includegraphics[width=\textwidth]{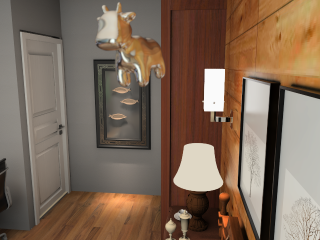}
  \end{subfigure}

  \vspace{0.0in}
  \begin{subfigure}[t]{\eximgwidth}
      \centering\includegraphics[width=\textwidth]{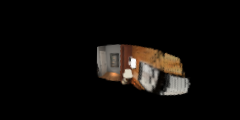}\caption{}
  \end{subfigure}
  \begin{subfigure}[t]{\eximgwidth}
      \centering\includegraphics[width=\textwidth]{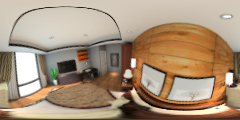}\caption{$\infty$ dB,\metricspace $0.00^\circ$}
  \end{subfigure}
  \begin{subfigure}[t]{\eximgwidth}
      \centering\includegraphics[width=\textwidth]{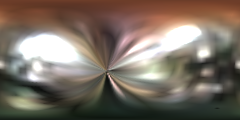}\caption{$10.44$ dB,\metricspace $11.60^\circ$}
  \end{subfigure}
  \begin{subfigure}[t]{\eximgwidth}
      \centering\includegraphics[width=\textwidth]{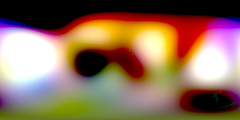}\caption{$8.95$ dB,\metricspace $17.46^\circ$}
  \end{subfigure}
  \begin{subfigure}[t]{\eximgwidth}
      \centering\includegraphics[width=\textwidth]{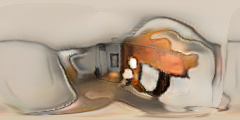}\caption{$15.86$ dB,\metricspace $5.43^\circ$}
  \end{subfigure}
  \begin{subfigure}[t]{\eximgwidth}
      \centering\includegraphics[width=\textwidth]{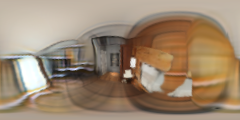}\caption{$16.72$ dB,\metricspace $4.61^\circ$}
  \end{subfigure}

  \vspace{0.0in}
  \begin{subfigure}[t]{\eximgwidth}
      \centering\includegraphics[width=\textwidth]{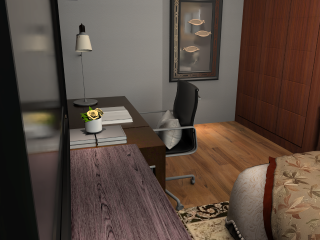}
  \end{subfigure}
  \begin{subfigure}[t]{\eximgwidth}
      \centering\includegraphics[width=\textwidth]{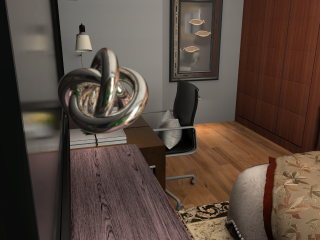}
  \end{subfigure}
  \begin{subfigure}[t]{\eximgwidth}
      \centering\includegraphics[width=\textwidth]{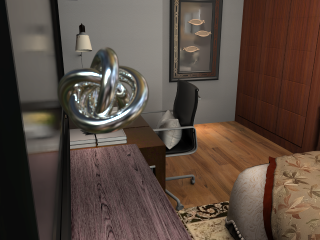}
  \end{subfigure}
  \begin{subfigure}[t]{\eximgwidth}
      \centering\includegraphics[width=\textwidth]{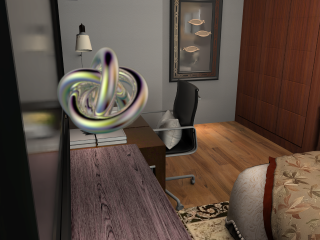}
  \end{subfigure}
  \begin{subfigure}[t]{\eximgwidth}
      \centering\includegraphics[width=\textwidth]{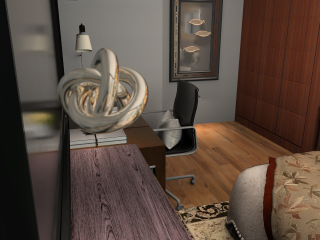}
  \end{subfigure}
  \begin{subfigure}[t]{\eximgwidth}
      \centering\includegraphics[width=\textwidth]{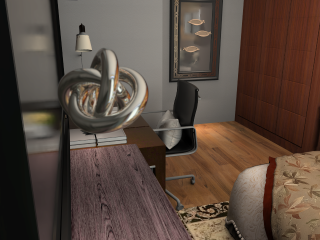}
  \end{subfigure}

  \vspace{0.0in}
  \begin{subfigure}[t]{\eximgwidth}
      \centering\includegraphics[width=\textwidth]{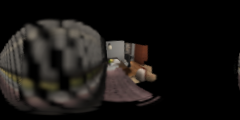}\caption{}
  \end{subfigure}
  \begin{subfigure}[t]{\eximgwidth}
      \centering\includegraphics[width=\textwidth]{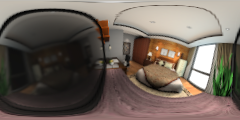}\caption{$\infty$ dB,\metricspace $0.00^\circ$}
  \end{subfigure}
  \begin{subfigure}[t]{\eximgwidth}
      \centering\includegraphics[width=\textwidth]{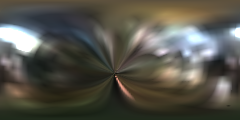}\caption{$12.92$ dB,\metricspace $7.61^\circ$}
  \end{subfigure}
  \begin{subfigure}[t]{\eximgwidth}
      \centering\includegraphics[width=\textwidth]{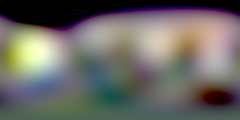}\caption{$11.41$ dB,\metricspace $10.16^\circ$}
  \end{subfigure}
  \begin{subfigure}[t]{\eximgwidth}
      \centering\includegraphics[width=\textwidth]{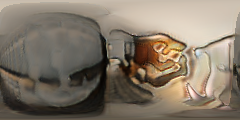}\caption{$16.65$ dB,\metricspace $4.08^\circ$}
  \end{subfigure}
  \begin{subfigure}[t]{\eximgwidth}
      \centering\includegraphics[width=\textwidth]{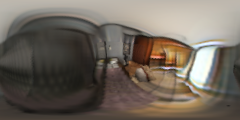}\caption{$19.79$ dB,\metricspace $3.07^\circ$}
  \end{subfigure}

  \vspace{0.0in}
  \begin{subfigure}[t]{\eximgwidth}
      \centering\includegraphics[width=\textwidth]{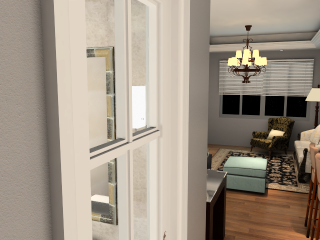}
  \end{subfigure}
  \begin{subfigure}[t]{\eximgwidth}
      \centering\includegraphics[width=\textwidth]{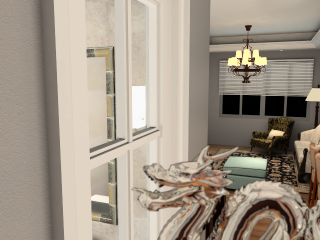}
  \end{subfigure}
  \begin{subfigure}[t]{\eximgwidth}
      \centering\includegraphics[width=\textwidth]{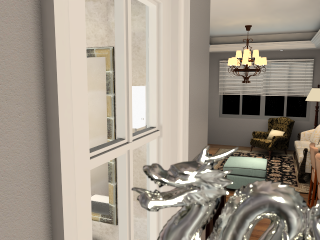}
  \end{subfigure}
  \begin{subfigure}[t]{\eximgwidth}
      \centering\includegraphics[width=\textwidth]{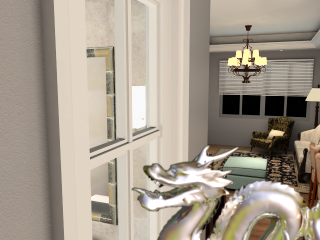}
  \end{subfigure}
  \begin{subfigure}[t]{\eximgwidth}
      \centering\includegraphics[width=\textwidth]{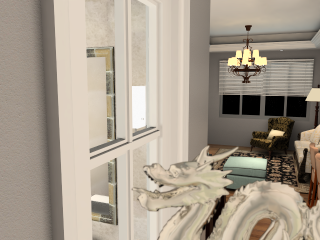}
  \end{subfigure}
  \begin{subfigure}[t]{\eximgwidth}
      \centering\includegraphics[width=\textwidth]{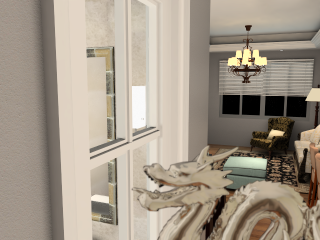}
  \end{subfigure}

  \vspace{0.0in}
  \begin{subfigure}[t]{\eximgwidth}
      \centering\includegraphics[width=\textwidth]{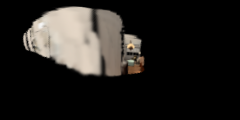}\caption{}
  \end{subfigure}
  \begin{subfigure}[t]{\eximgwidth}
      \centering\includegraphics[width=\textwidth]{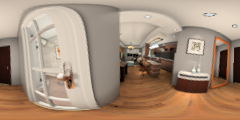}\caption{$\infty$ dB,\metricspace $0.00^\circ$}
  \end{subfigure}
  \begin{subfigure}[t]{\eximgwidth}
      \centering\includegraphics[width=\textwidth]{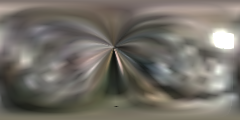}\caption{$13.71$ dB,\metricspace $6.92^\circ$}
  \end{subfigure}
  \begin{subfigure}[t]{\eximgwidth}
      \centering\includegraphics[width=\textwidth]{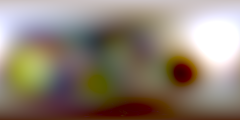}\caption{$13.23$ dB,\metricspace $9.46^\circ$}
  \end{subfigure}
  \begin{subfigure}[t]{\eximgwidth}
      \centering\includegraphics[width=\textwidth]{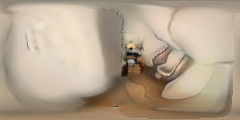}\caption{$16.61$ dB,\metricspace $4.50^\circ$}
  \end{subfigure}
  \begin{subfigure}[t]{\eximgwidth}
      \centering\includegraphics[width=\textwidth]{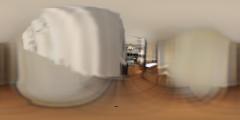}\caption{$19.06$ dB,\metricspace $2.93^\circ$}
  \end{subfigure}

    \vspace{-3mm}
        \caption{Estimated environment maps and images with inserted relit virtual objects for scenes from our synthetic InteriorNet test set. The leftmost column displays the input reference image (our method also takes a second input image) and the portion of the environment map that is visible in this reference image (used to visualize the portion of the environment map which must be hallucinated in black). We display quantitative metrics (PSNR and RGB Angular Error) for the predicted environment maps below each method's results. Our method outperforms all competing methods, both qualitatively and quantitatively, producing realistic environment maps with plausible unobserved regions. Inserted specular virtual objects relit with our environment maps have highlights and reflected colors that are closer to ground truth than those relit by baseline methods.}
    \label{fig:comparisons}
  
\end{figure*}

\begin{figure}
\begin{center}
\newcommand{\width}{\columnwidth}
\includegraphics[width=\width]{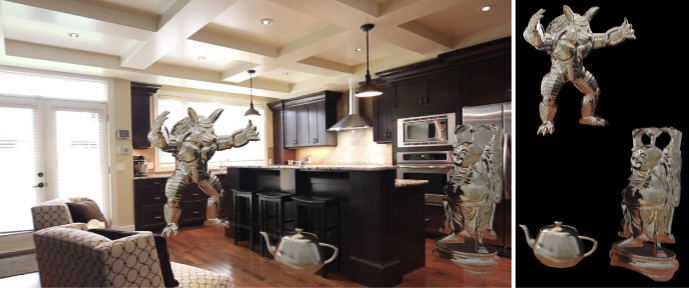}
\includegraphics[width=\width]{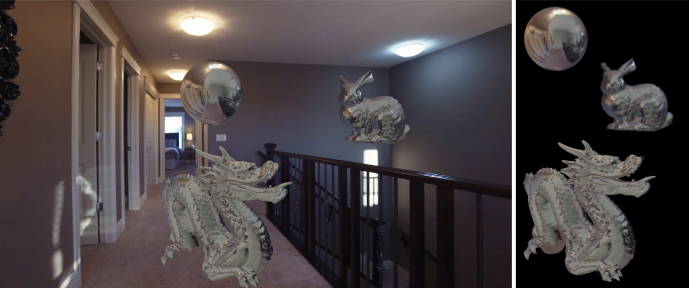}
\includegraphics[width=\width]{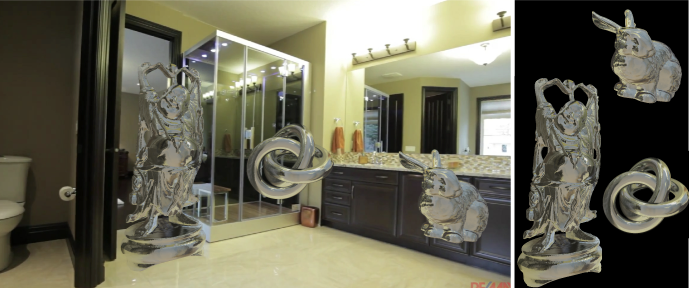}
\caption{Real images from the RealEstate10K dataset~\cite{zhou2018stereo} with inserted virtual objects relit with spatially-varying lighting estimated by our method. As in Fig.~\ref{fig:teaser}, we render perfectly specular objects by querying our multiscale lighting volume for spatially varying lighting values. The reflected colors are consistent with the scene content in the original image. Please see our supplemental video for examples of relit objects moving through the scene.}
\label{fig:realestate_insertion}
\end{center}
\end{figure}

\newcommand{\epimapwidth}{.16\textwidth}
\newcommand{\riwidth}{.135\textwidth}

\begin{figure}
\captionsetup[subfigure]{font=scriptsize,labelformat=empty,aboveskip=1pt,belowskip=2pt}
  \centering

  \begin{subfigure}[t]{\riwidth}
      \centering Ref. Image
  \end{subfigure}
  \begin{subfigure}[t]{\epimapwidth}
      \centering Neural Illum.~\cite{song2019neural}
  \end{subfigure}
  \begin{subfigure}[t]{\epimapwidth}
      \centering Ours
  \end{subfigure}

  \vspace{0.0in}
  \begin{subfigure}[t]{\riwidth}
      \centering\includegraphics[width=\textwidth]{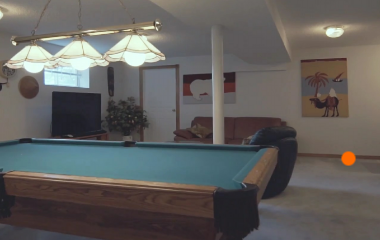}
  \end{subfigure}
  \begin{subfigure}[t]{\epimapwidth}
      \centering\includegraphics[width=\textwidth]{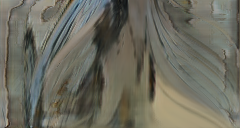}
  \end{subfigure}
  \begin{subfigure}[t]{\epimapwidth}
      \centering\includegraphics[width=\textwidth]{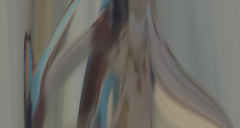}
  \end{subfigure}

        \caption{
        Our estimated illumination is more spatially-coherent than that of Neural Illumination~\cite{song2019neural}. Each row in the right two images is a row from environment maps rendered along the camera ray marked in orange (depth increases with lower rows). Our estimated illumination varies much more smoothly as a function of 3D position.}
    \label{fig:epipolar}

\end{figure}

\subsection{Comparisons to baseline methods}

We compare our method to trained models of DeepLight~\cite{legendre2019deeplight} and Garon \etal~\cite{garon2019} (which both take a single image as input) provided by the authors, and to a generous re-implementation of Neural Illumination~\cite{song2019neural} that has access to ground-truth geometry, in order to provide a fair comparison against our method which requires stereo input. Note that all quantitative metrics are computed on LDR images, due to the limitations of the InteriorNet dataset.

DeepLight~\cite{legendre2019deeplight} takes in a single image and outputs one HDR lighting map for the entire scene. The training data used for supervision is a set of three light probes placed 60cm in front of the camera; thus, this location is where the predicted lighting should be most accurate. The lighting is output in the form of a $32 \times 32$ HDR image of a mirror ball light probe. In order to compare these results with our method, we resample the mirror ball onto a higher resolution $120\times 240$ spherical environment map, rotated to match the orientation of the target environment map.
Because DeepLight does not predict spatially varying lighting, it underperforms our method on our test set of environment maps at various locations within the scene (see Table~\ref{tab:comparisons_both}). Qualitatively, the limitations of a single environment map are apparent in relighting results since inserted objects cannot be correctly relit as they are moved around the scene, as we show in our supplementary materials.

Garon \etal~\cite{garon2019} predicts spatially-varying lighting from a single image. Given a particular pixel on an object surface, the method is supervised to match the lighting 10cm away from that surface in the normal direction.
This allows for some spatially-varying effects but heavily restricts the supported lighting locations. Additionally, this method predicts a low-dimensional representation of the environment map as 36 spherical harmonic coefficients for efficiency.
Since our test set consists of higher resolution $120\times 240$ environment maps sampled at locations that are not restricted to be near surfaces, this method performs significantly worse in our quantitative comparisons (see Table~\ref{tab:comparisons_both}). Qualitatively, this lighting representation is sufficient for relighting diffuse objects, but its low resolution prevents it from plausibly relighting glossy and specular objects (see Fig.~\ref{fig:comparisons}).

Neural Illumination~\cite{song2019neural} estimates lighting at any 3D scene point by first predicting a per-pixel geometry for a single input image, warping the input image with this geometry to render an incomplete environment map, using a 2D CNN to inpaint unobserved content, and finally using another 2D CNN to convert the environment map to HDR. We did not have access to the authors' original implementation, so we implemented a generous baseline that uses the \textbf{ground truth} depth to warp the visible scene content into an incomplete spherical environment map, then uses a 2D CNN to complete the lighting map (see the supplementary PDF for additional details).
Despite having access to the ground truth geometry for the observed portion of the scene, this method is not guaranteed to produce spatially-coherent lighting predictions for the unobserved regions of a given scene, because the 2D completion CNN is run \emph{independently} at each queried location. In contrast, our single multiscale 3D representation of the entire scene guarantees consistency across different 3D lighting locations. Figure~\ref{fig:epipolar} demonstrates how our method produces spatially-coherent lighting estimations that vary much more smoothly with 3D position than lighting estimated by Neural Illumination. Furthermore, Neural Illumination contains less realistic hallucinations of unobserved scene content, as shown in Fig.~\ref{fig:comparisons}. This is because it is much harder for their 2D completion CNN to learn meaningful implicit priors on environment maps since they can be observed with arbitrary rotations. In contrast, our strategy predicts lighting in a canonical 3D frame instead of in a 2D environment map pixel space, so it is much easier for our network to learn meaningful priors on the distribution of 3D lighting.

Please see our supplementary video and PDF for additional comparisons of relit inserted objects that showcase the temporal consistency and realistic spatially-varying illumination produced by our method.

\subsection{Comparisons to ablations of our method}

We also present quantitative results from two ablations of our method in Table~\ref{tab:comparisons_both}.
Our ``MPI only'' ablation only uses our prediction of observed scene geometry to render the spherical environment map (and fills unobserved regions with grey). In this case, the network cannot add light values to the unseen parts of the scene, so the resulting environment maps are largely incomplete. Since these missing regions are the most important for relighting objects inserted into the scene, we see a significant decrease in quality. Interestingly, our full method even outperforms the ``MPI only'' ablation for the observed content, which shows that our multiscale volume completion network learns to correct errors in MPI prediction. Our ``No $\lossfun{adv}$'' ablation omits the adversarial loss when training the volume completion network. The resulting environment maps are slightly better quantitatively, but contain less high frequency detail, resulting in less realistic appearance when rendering glossy inserted objects. Please see our supplementary materials for qualitative results from these ablations.

\section{Discussion}

This paper demonstrates that using a persistent 3D lighting model of the scene is a compelling strategy for estimating spatially-coherent illumination from images. We have chosen a multiscale volumetric lighting representation to make this approach tractable and proposed a deep learning pipeline to predict this lighting representation using only images as supervision. Our results demonstrate that this strategy produces plausible spatially-coherent lighting and outperforms prior state-of-the-art work.

However, we have just touched the surface of possible 3D lighting representations for this task. An exciting direction would be to develop models that adaptively allocate 3D samples as needed to represent a scene, rather than being limited to a fixed multiresolution sampling pattern. We hope that this work enables future progress in predicting 3D scene representations for lighting estimation and other inverse rendering tasks.

{\small
\bibliographystyle{ieee_fullname}
\bibliography{paper}
}
\clearpage
\appendix

\section{Supplementary Video}

We encourage readers to view our included supplementary video for a brief overview of our method and qualitative comparisons between our method and baselines that showcase our method's spatial coherence by inserting specular virtual objects that move along smooth paths.

\section{Multiscale Lighting Volume Details}

Our multiscale lighting volume consists of 5 scales of $64 \times 64 \times 64$ \RGBA volumes. As illustrated in Figure 3 in the main paper, each scale's volume has half the side length of the previous scale's volume.

\begin{figure}
\captionsetup[subfigure]{aboveskip=3pt,belowskip=3pt}
\begin{center}
\newcommand{\width}{\columnwidth}

  \begin{subfigure}[t]{\columnwidth}
      \centering 
    \includegraphics[width=\width]{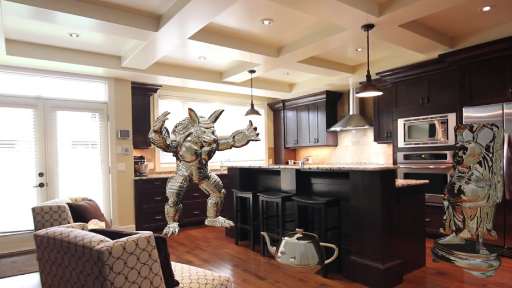} 
    \caption{Single environment map per object}
    \label{sub:nospat}
  \end{subfigure}
  \begin{subfigure}[t]{\columnwidth}
      \centering 
    \includegraphics[width=\width]{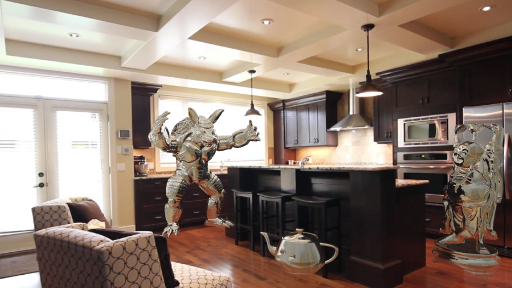}
    \caption{Fully spatially varying lighting}
    \label{sub:spat}
  \end{subfigure}
\caption{Visualization of our method's ability to realistically render spatially-varying lighting across each object's surface. In (a), we use a single environment map predicted by our method at the object's centroid to illuminate the entire object. In (b), we illuminate each point on each object's surface by tracing rays through our predicted volume, so each point is effectively illuminated by a different environment map. This results in more realistic lighting effects, as can be seen in the reflection of the floor in the armadillo's legs and the Buddha's statue's base and the reflection of the window in the Buddha statue's face. This difference is less pronounced in smaller objects such as the teapot.}
\label{fig:spatvary}
\end{center}
\end{figure}

\section{Illumination Rendering Details}

For training, we implement the volume rendering procedure described in the main paper by sampling each volume in our multiscale lighting representation on a set of concentric spheres around the target environment map location with trilinear interpolation. We use 128 spheres per scale, sampled evenly in radius between the closest and furthest voxels from the target location. Then, we alpha-composite these spheres ($128 \times 5=640$ total spheres) from outermost to innermost to render the environment map at that location.

For fast relighting performance at test time, we implement the volume rendering as standard ray tracing using CUDA. We intersect each camera ray with each virtual object, and then trace rays through the volume from that intersection location to compute the incident illumination to correctly shade that pixel. This means that the illumination we use for relighting virtual objects varies spatially both between different objects as well as across the geometry of each object. This effect is quite difficult to simulate with prior lighting estimation work such as Neural Illumination~\cite{song2019neural}, since this would require running a deep network tens of thousands of times to predict an environment map for each camera ray intersecting the virtual object. In contrast, our method only requires one pass of network inference to predict a multiscale volumetric lighting estimation, and full spatially-varying lighting across inserted objects is then handled by ray tracing through our volumes. 

Figure~\ref{fig:spatvary} illustrates how our spatially-varying lighting across the surfaces of inserted objects adds realism. In the top image, we render all points on each object's surface using a single environment map predicted by our method at the object's centroid. In the bottom image, we render each point by ray tracing through our predicted lighting volume, so each point is effectively illuminated by a different environment map. We can see that the inserted virtual objects have a more realistic appearance in the bottom image. For example, the armadillo's legs and the Buddha statue's base correctly reflect the floor while the Buddha statue's face correctly reflects the windows instead of the black counter.

\section{Network Architectures}

\paragraph{MPI prediction network (Table~\ref{table:arch1})}

Our MPI prediction network is a 3D encoder-decoder CNN with skip connections. We use residual blocks~\cite{he16} with layer normalization~\cite{ba16} for all layers, strided convolutions for downsampling within the encoder, and nearest-neighbor upsampling within the decoder. This network outputs a 3D array of \RGBA values and a scalar 3D array of blending weights between 0 and 1. 

We compute a ``background'' image as the average RGB over all depth planes in the output array, and use the ``background + blending weights'' MPI parameterization~\cite{zhou2018stereo}, where each MPI plane's RGB colors are defined as a convex combination of the input reference image and the predicted ``background'' image, using the predicted blending weights. 

\paragraph{Volume completion network (Table~\ref{table:arch1})}

We use the same 3D encoder-decoder CNN architecture detailed above for all 5 scales of our volume completion network, with separate weights per scale. At each scale, the network predicts an \RGBA volume as well as a scalar volume of blending weights between 0 and 1. We parameterize each scale's output as a convex combination of the input resampled volume at that scale and the network's output \RGBA volume, using the network's predicted blending weights.

\paragraph{Discriminator network (Table~\ref{table:arch2})}

We use a 2D CNN PatchGAN~\cite{isola17} architecture with spectral normalization~\cite{miyato18}. 

\begin{table}[t]
\begin{center}
\resizebox{3.25in}{!}{

\begin{tabular}{ c | l | c c c c}

\hline \multicolumn{3}{c}{\textbf{Encoder}}\\ \hline
 1 & $3\times 3 \times 3$ conv, 8 features &
 $ H \times W \times D \times 8 $ \\ 
 
 2 & $3\times 3 \times 3$ conv, 16 features, stride 2 &
 $ H/2 \times W/2 \times D/2 \times 16 $ \\ 

 3-4 & ($3\times 3 \times 3$ conv, 16 features) $\times 2$, residual  &
 $ H/2 \times W/2 \times D/2 \times 16 $ \\ 
 
 5 & $3\times 3 \times 3$ conv, 32 features, stride 2 &
 $ H/4 \times W/4 \times D/4 \times 32 $ \\ 
 
 6-7 & ($3\times 3 \times 3$ conv, 32 features) $\times 2$, residual  &
 $ H/4 \times W/4 \times D/4 \times 32 $ \\ 
 
 8 & $3\times 3 \times 3$ conv, 64 features, stride 2 &
 $ H/8 \times W/8 \times D/8 \times 64 $ \\ 
 
 9-10 & ($3\times 3 \times 3$ conv, 32 features) $\times 2$, residual  &
 $ H/8 \times W/8 \times D/8 \times 64 $ \\ 
 
 11 & $3\times 3 \times 3$ conv, 128 features, stride 2 &
 $ H/16 \times W/16 \times D/16 \times 128 $ \\ 
 
 12-13 & ($3\times 3 \times 3$ conv, 32 features) $\times 2$, residual  &
 $ H/16 \times W/16 \times D/16 \times 128 $ \\ 
 
 \hline \multicolumn{3}{c}{\textbf{Decoder}}\\ \hline
 
 14 & $2\times$ nearest neighbor upsample &
 $ H/8 \times W/8 \times D/8 \times 128 $ \\ 
 
 15 & concatenate 14 and 10 &
 $ H/8 \times W/8 \times D/8 \times (128+64) $ \\
 
 16 & $3\times 3 \times 3$ conv, 64 features &
 $ H/8 \times W/8 \times D/8 \times 64 $ \\
 
 17-18 & ($3\times 3 \times 3$ conv, 64 features) $\times 2$, residual  &
 $ H/8 \times W/8 \times D/8 \times 64 $ \\ 
 
 19 & $2\times$ nearest neighbor upsample &
 $ H/4 \times W/4 \times D/4 \times 64 $ \\ 
 
 20 & concatenate 19 and 7 &
 $ H/4 \times W/4 \times D/4 \times (64+32) $ \\
 
 21 & $3\times 3 \times 3$ conv, 32 features &
 $ H/4 \times W/4 \times D/4 \times 32 $ \\
 
 22-23 & ($3\times 3 \times 3$ conv, 32 features) $\times 2$, residual  &
 $ H/4 \times W/4 \times D/4 \times 32 $ \\ 
 
 24 & $2\times$ nearest neighbor upsample &
 $ H/2 \times W/2 \times D/2 \times 32 $ \\ 
 
 25 & concatenate 24 and 4 &
 $ H/2 \times W/2 \times D/2 \times (32+16) $ \\
 
 26 & $3\times 3 \times 3$ conv, 16 features &
 $ H/2 \times W/2 \times D/2 \times 16 $ \\
 
 27-28 & ($3\times 3 \times 3$ conv, 16 features) $\times 2$, residual  &
 $ H/2 \times W/2 \times D/2 \times 16 $ \\ 
 
 29 & $2\times$ nearest neighbor upsample &
 $ H \times W \times D \times 16 $ \\ 
 
 30 & concatenate 29 and 1 &
 $ H \times W \times D \times (32+16) $ \\
 
 26 & $3\times 3 \times 3$ conv, 16 features &
 $ H \times W \times D \times 16 $ \\
 
 27 & $3\times 3 \times 3$ conv, 5 features (sigmoid) &
 $ H \times W \times D \times 5 $ \\ 

\end{tabular}
}
\vspace{0.02in}
\caption{\textbf{3D CNN network architecture used for MPI prediction and volume completion networks.} All convolutional layers use a ReLu activation, except for the final layer which uses a sigmoid activation.
\label{table:arch1}}
\end{center}
\vspace{-0.1in}
\end{table}

\begin{table}[t]
\begin{center}
\resizebox{3.25in}{!}{

\begin{tabular}{ c | l | c c c c}

\hline \multicolumn{3}{c}{\textbf{Encoder}}\\ \hline
 1 & $4\times 4$ conv, 64 features, stride 2 &
 $ H/2 \times W/2 \times D/2 \times 64 $ \\ 
 
 2 & $4\times 4$ conv, 128 features, stride 2 &
 $ H/4 \times W/4 \times D/4 \times 128 $ \\ 
 
 3 & $4\times 4$ conv, 256 features, stride 2 &
 $ H/8 \times W/8 \times D/8 \times 256 $ \\ 
 
 4 & $4\times 4$ conv, 512 features, stride 2 &
 $ H/16 \times W/16 \times D/16 \times 512 $ \\ 
 
 5 & $4\times 4$ conv, 1 feature &
 $ H/16 \times W/16 \times D/16 \times 1 $ \\ 

\end{tabular}
}
\vspace{0.02in}
\caption{\textbf{2D CNN discriminator network architecture.} All convolutional layers use a Leaky ReLu activation with $\alpha=0.2$, except for the final layer.
\label{table:arch2}}
\end{center}
\vspace{-0.1in}
\end{table}

\section{Baseline Method Details}

DeepLight~\cite{legendre2019deeplight} and Garon \etal~\cite{garon2019} output HDR environment maps with unknown scales, since camera exposure is a free parameter. For fair comparisons, we scale their environment maps so that their average radiance matches the average radiance of the ground-truth environment maps for our quantitative and qualitative results when using the InteriorNet dataset. There is no ground-truth environment map for our real photograph results, so we scale their predicted environment map so that their average radiance matches the average radiance of the reference image.

 \renewcommand{\eximgwidth}{.19\textwidth}
\newcommand{\envmapwidth}{.31\textwidth}

\begin{figure*}
\captionsetup[subfigure]{font=scriptsize,labelformat=empty,aboveskip=1pt,belowskip=2pt}
  \centering

  \begin{subfigure}[t]{\eximgwidth}
      \centering Ref. Image
  \end{subfigure}
  \begin{subfigure}[t]{\eximgwidth}
      \centering DeepLight~\cite{legendre2019deeplight}
  \end{subfigure}
  \begin{subfigure}[t]{\eximgwidth}
      \centering Garon \etal~\cite{garon2019}
  \end{subfigure}
  \begin{subfigure}[t]{\eximgwidth}
      \centering Neural Illum.~\cite{song2019neural}
  \end{subfigure}
  \begin{subfigure}[t]{\eximgwidth}
      \centering Ours
  \end{subfigure}

  \vspace{0.0in}
  \begin{subfigure}[t]{\eximgwidth}
      \centering\includegraphics[width=\textwidth]{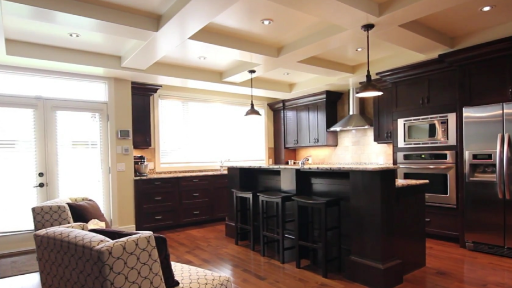}
  \end{subfigure}
  \begin{subfigure}[t]{\eximgwidth}
      \centering\includegraphics[width=\textwidth]{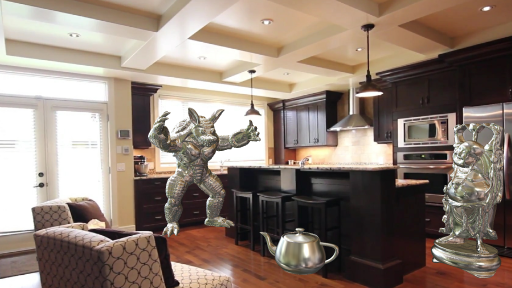}
	\centering\includegraphics[width=\envmapwidth]{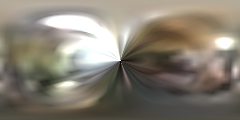}
	\centering\includegraphics[width=\envmapwidth]{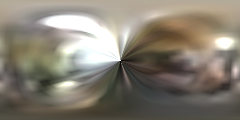}
	\centering\includegraphics[width=\envmapwidth]{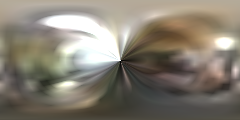}
  \end{subfigure}
  \begin{subfigure}[t]{\eximgwidth}
      \centering\includegraphics[width=\textwidth]{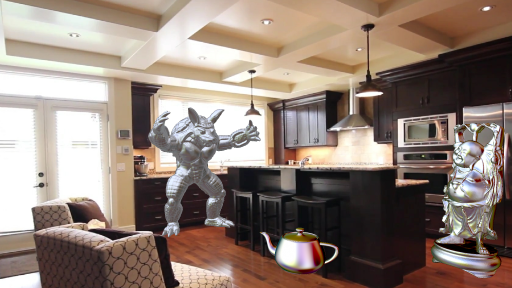}
	\centering\includegraphics[width=\envmapwidth]{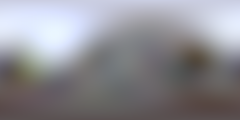}
	\centering\includegraphics[width=\envmapwidth]{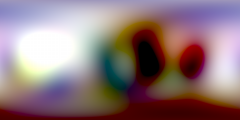}
	\centering\includegraphics[width=\envmapwidth]{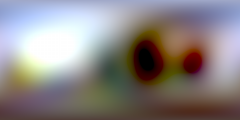}
  \end{subfigure}
  \begin{subfigure}[t]{\eximgwidth}
      \centering\includegraphics[width=\textwidth]{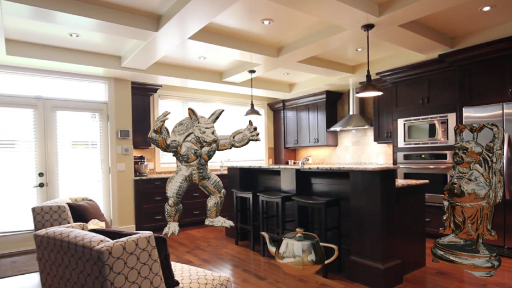}
	\centering\includegraphics[width=\envmapwidth]{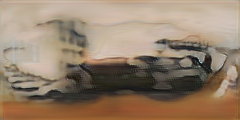}
	\centering\includegraphics[width=\envmapwidth]{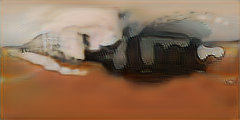}
	\centering\includegraphics[width=\envmapwidth]{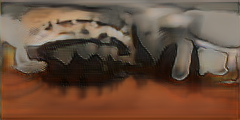}
  \end{subfigure}
  \begin{subfigure}[t]{\eximgwidth}
      \centering\includegraphics[width=\textwidth]{supplement_table/imgs/010_ours.png}
	\centering\includegraphics[width=\envmapwidth]{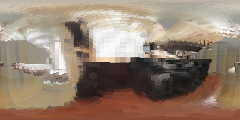}
	\centering\includegraphics[width=\envmapwidth]{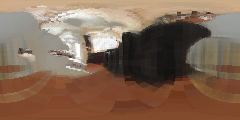}
	\centering\includegraphics[width=\envmapwidth]{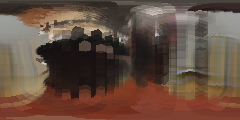}
  \end{subfigure}

  \vspace{0.0in}
  \begin{subfigure}[t]{\eximgwidth}
      \centering\includegraphics[width=\textwidth]{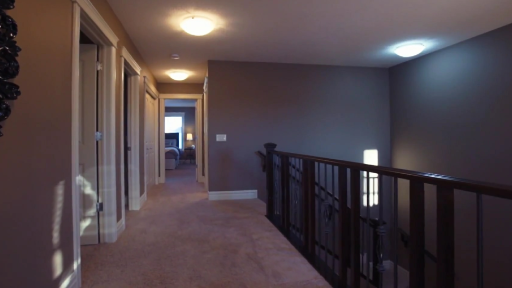}
  \end{subfigure}
  \begin{subfigure}[t]{\eximgwidth}
      \centering\includegraphics[width=\textwidth]{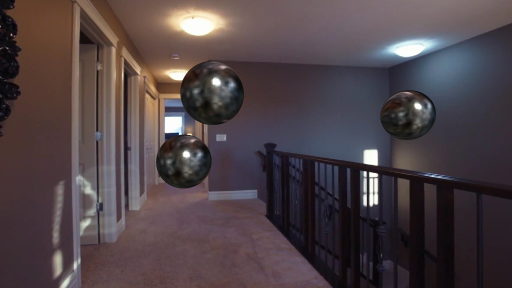}
	\centering\includegraphics[width=\envmapwidth]{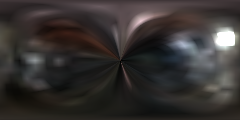}
	\centering\includegraphics[width=\envmapwidth]{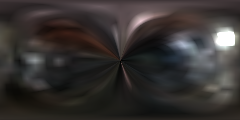}
	\centering\includegraphics[width=\envmapwidth]{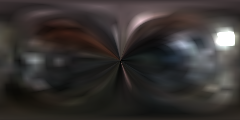}
  \end{subfigure}
  \begin{subfigure}[t]{\eximgwidth}
      \centering\includegraphics[width=\textwidth]{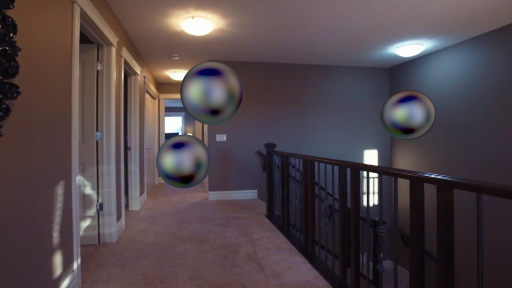}
	\centering\includegraphics[width=\envmapwidth]{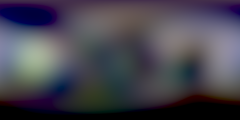}
	\centering\includegraphics[width=\envmapwidth]{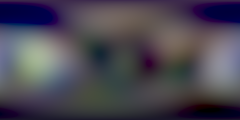}
	\centering\includegraphics[width=\envmapwidth]{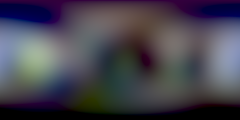}
  \end{subfigure}
  \begin{subfigure}[t]{\eximgwidth}
      \centering\includegraphics[width=\textwidth]{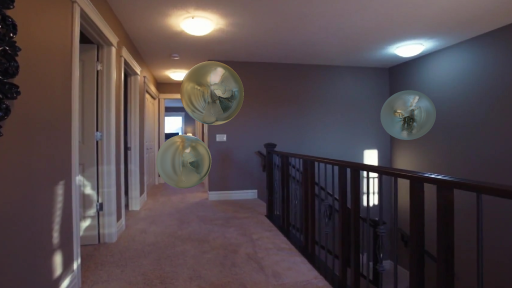}
	\centering\includegraphics[width=\envmapwidth]{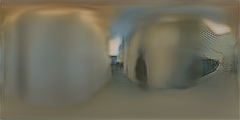}
	\centering\includegraphics[width=\envmapwidth]{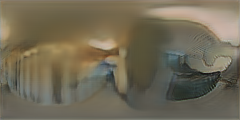}
	\centering\includegraphics[width=\envmapwidth]{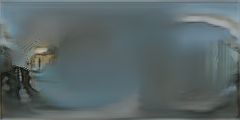}
  \end{subfigure}
  \begin{subfigure}[t]{\eximgwidth}
      \centering\includegraphics[width=\textwidth]{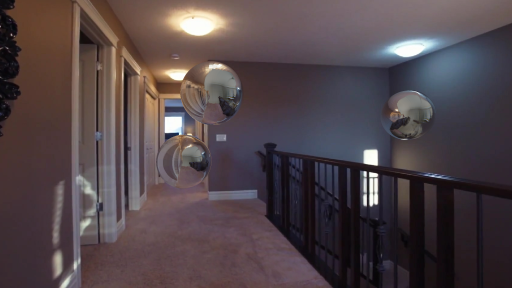}
	\centering\includegraphics[width=\envmapwidth]{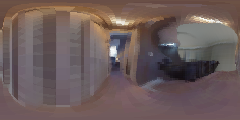}
	\centering\includegraphics[width=\envmapwidth]{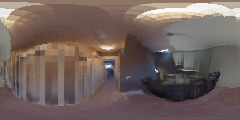}
	\centering\includegraphics[width=\envmapwidth]{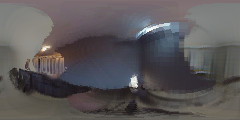}
  \end{subfigure}

  \vspace{0.0in}
  \begin{subfigure}[t]{\eximgwidth}
      \centering\includegraphics[width=\textwidth]{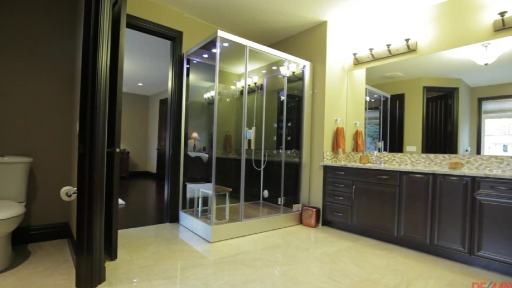}
  \end{subfigure}
  \begin{subfigure}[t]{\eximgwidth}
      \centering\includegraphics[width=\textwidth]{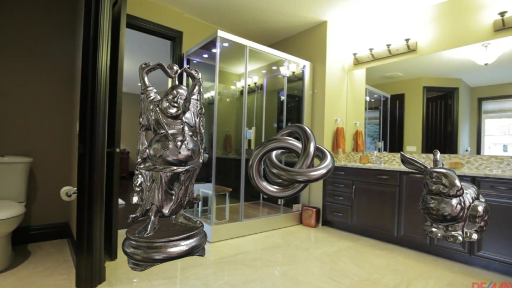}
	\centering\includegraphics[width=\envmapwidth]{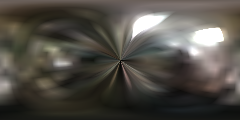}
	\centering\includegraphics[width=\envmapwidth]{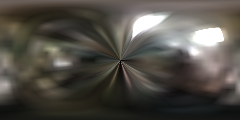}
	\centering\includegraphics[width=\envmapwidth]{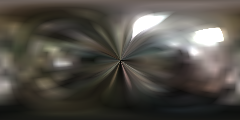}
  \end{subfigure}
  \begin{subfigure}[t]{\eximgwidth}
      \centering\includegraphics[width=\textwidth]{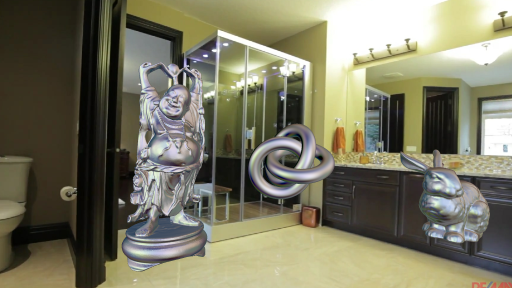}
	\centering\includegraphics[width=\envmapwidth]{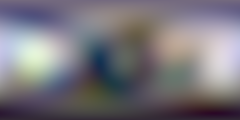}
	\centering\includegraphics[width=\envmapwidth]{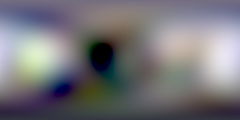}
	\centering\includegraphics[width=\envmapwidth]{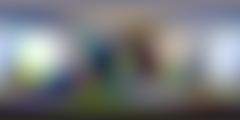}
  \end{subfigure}
  \begin{subfigure}[t]{\eximgwidth}
      \centering\includegraphics[width=\textwidth]{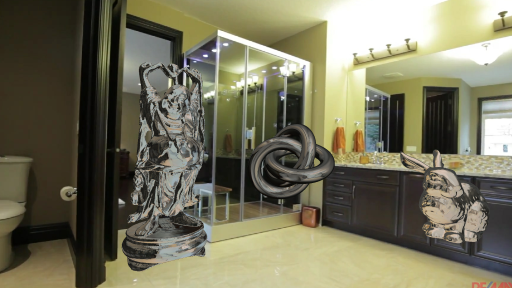}
	\centering\includegraphics[width=\envmapwidth]{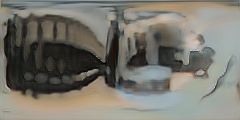}
	\centering\includegraphics[width=\envmapwidth]{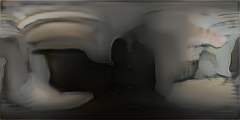}
	\centering\includegraphics[width=\envmapwidth]{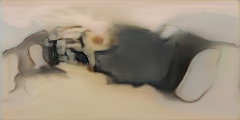}
  \end{subfigure}
  \begin{subfigure}[t]{\eximgwidth}
      \centering\includegraphics[width=\textwidth]{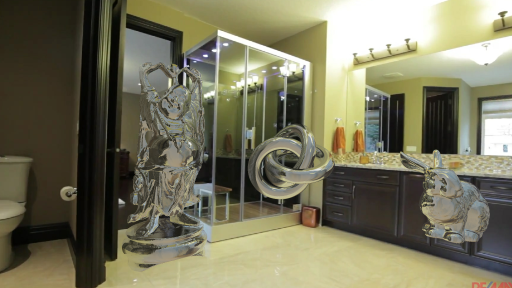}
	\centering\includegraphics[width=\envmapwidth]{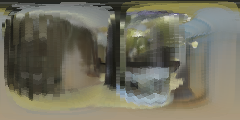}
	\centering\includegraphics[width=\envmapwidth]{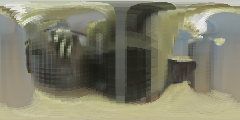}
	\centering\includegraphics[width=\envmapwidth]{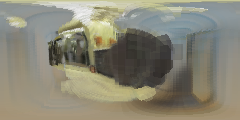}
  \end{subfigure}

    \vspace{-3mm}
        \caption{Qualitative comparison of real images from the RealEstate10K dataset~\cite{zhou2018stereo} with relit inserted virtual objects and corresponding environment maps. DeepLight~\cite{legendre2019deeplight} only estimates a single environment map for the entire scene, so virtual objects at different locations do not realistically reflect scene content. Garon \etal~\cite{garon2019} estimate a low-dimensional lighting representation at each pixel, so their lighting does not realistically vary across 3D locations along the same camera ray. Furthermore, their low-dimensional representation does not contain sufficient high frequency detail for rendering specular objects, so inserted objects have a more diffuse appearance. Neural Illumination~\cite{song2019neural} has trouble correctly preserving scene content colors from the input image. Additionally, their method separately predicts unobserved content for the environment map at each 3D location so their predicted lighting is not as spatially-coherent. Our results contain more plausible spatially-coherent reflections, and we can see that the  colors of virtual objects in our results are more consistent with the scene content in the original image.}
    \label{fig:supp_comparisons}
  
\end{figure*}

Neural Illumination~\cite{song2019neural} does not have an available implementation, so we implement and train a generous baseline version of their method. Their published method trains a 2D CNN network to predict per-pixel geometry from a single input image, uses this geometry to warp input image pixels into the target environment map, trains another 2D CNN to complete the unobserved areas of this environment map, and trains a final 2D CNN to convert this environment map to HDR. To enable a generous comparison with our method, which uses a stereo pair of images as input, we remove the first CNN from the Neural Illumination method, and instead use the ground-truth depth to reproject input image pixels into the target environment map for all quantitative and qualitative comparisons on our InteriorNet test set (we use our method's estimated MPI geometry for qualitative results on the RealEstate dataset where no ground truth is available). Additionally, since we assume InteriorNet renderings are captured with a known invertible tone-mapping function, we omit Neural Illumination's LDR to HDR conversion network and instead just apply the exact inverse tone-mapping function to their completed LDR environment maps. For a fair comparison with our method's results, we use the same architecture as Table~\ref{table:arch1} for the environment map completion network, but with 2D kernels instead of 3D, and the same discriminator as Table~\ref{table:arch2}. We train this generous Neural Illumination baseline on the same InteriorNet training dataset that we use to train our method.

\section{Additional Results}

Figure~\ref{fig:supp_comparisons} contains additional qualitative results for specular virtual objects relit with baseline methods and our algorithm. We can see that our results are more spatially-coherent and contain realistic reflections of scene content.



\end{document}